\documentclass[runningheads]{llncs}
\pdfoutput=1
\usepackage{graphicx}
\usepackage[dvipsnames]{xcolor}
\usepackage[colorlinks=true]{hyperref}
\hypersetup{citecolor=SeaGreen}
\usepackage{amssymb}
\usepackage{booktabs}
\usepackage{adjustbox}
\usepackage{multirow,rotating}

\begin{document}
\title{Anatomy-Aware Self-supervised Fetal MRI Synthesis from Unpaired Ultrasound Images}
\titlerunning{Anatomy-Aware Self-supervised Fetal MRI Synthesis from Unpaired US}
\author{Jianbo Jiao \inst{1},
Ana I.L. Namburete \inst{1},
Aris T. Papageorghiou \inst{2},
J. Alison Noble \inst{1}}
\authorrunning{J. Jiao et al.}
\institute{Department of Engineering Science, University of Oxford, Oxford, UK \email{jianbo.jiao@eng.ox.ac.uk} \and
Nuffield Department of Women’s \& Reproductive Health, University of Oxford, UK
}
\maketitle              
\begin{abstract}
  Fetal brain magnetic resonance imaging (MRI) offers exquisi\-te images of the developing brain but is not suitable for anomaly screening. For this ultrasound (US) is employed. While expert sonographers are adept at reading US images, MR images are much easier for non-experts to interpret. Hence in this paper we seek to produce images with MRI-like appearance directly from clinical US images. Our own clinical motivation is to seek a way to communicate US findings to patients or clinical professionals unfamiliar with US, but in medical image analysis such a capability is potentially useful, for instance, for US-MRI registration or fusion. Our model is self-supervised and end-to-end trainable. Specifically, based on an assumption that the US and MRI data share a similar anatomical latent space, we first utilise an extractor to determine shared latent features, which are then used for data synthesis. Since paired data was unavailable for our study (and rare in practice), we propose to enforce the distributions to be similar instead of employing pixel-wise constraints, by adversarial learning in both the image domain and latent space. Furthermore, we propose an adversarial structural constraint to regularise the anatomical structures between the two modalities during the synthesis. A cross-modal attention scheme is proposed to leverage non-local spatial correlations. The feasibility of the approach to produce realistic looking MR images is demonstrated quantitatively and with a qualitative evaluation compared to real fetal MR images.
\end{abstract}
\section{Introduction}
Ultrasound (US) imaging is widely employed in image-based diagnosis, as it is portable, real-time, and safe for body tissue. Obstetric US is the most commonly used clinical imaging technique to monitor fetal development. Clinicians use fetal brain US imaging (fetal neurosonography) to detect abnormalities in the fetal brain and growth restriction. However, fetal neurosonography suffers from acoustic shadows and occlusions caused by the fetal skull. MRI is unaffected by the presence of bone and typically provides good and more complete spatial detail of the full anatomy~\cite{pugash2008prenatal}. On the other hand, MRI is time-consuming and costly, making it unsuitable for fetal anomaly screening, but it is often used for routine fetal brain imaging in the second and third trimester~\cite{bulas2013benefits}.
Therefore, we seek to generate MR images of fetal brains directly from clinical US images.

Medical image synthesis has received growing interest in recent years.
Most prior work has focused on the synthesis of MR/CT (computed tomography) images~\cite{zhao2018towards,nie2017medical,yang2018unpaired} or retinal images~\cite{costa2018end}. Some works simulate US for image alignment~\cite{kuklisova2013registration,king2010registering}.
Prior to the deep learning era, medical image synthesis was primarily based on segmentation and atlases. Taking MR-to-CT image synthesis as an example, in segmentation-based approaches~\cite{berker2012mri,delpon2016comparison}, different tissue classes are segmented for the MR image, followed by an intensity-filling step to generate the corresponding CT image. Atlas-based methods~\cite{sjolund2015generating,catana2010towards} first register an MR atlas to the input MRI, and then apply the transformation to synthesise the corresponding CT image from a CT atlas. However, these methods highly depend on the accuracy of segmentation and registration. With the popularity of deep learning techniques, recent convolutional neural network (CNN) based methods have achieved promising results for image synthesis. Some works~\cite{roy2017synthesizing,nie2017medical,zhao2018towards} have directly learned the mapping from MR to CT via a CNN architecture, assuming a large number of MR-CT data pairs. To overcome a paired data requirement, other approaches~\cite{yang2018unpaired,zhang2018translating} have utilised a CycleGAN architecture~\cite{CycleGAN2017} for image-to-image translation. Although they do not need perfectly registered data, previous methods have either needed weakly paired data (from the same subject) or supervision from other auxiliary tasks (segmentation). In addition, the aforementioned works do not consider synthesis from US images, which is much more challenging than anatomical imaging (CT/MR) due to its more complex image formation process.

\begin{figure}[t]
  \centering
  \includegraphics[width=\textwidth]{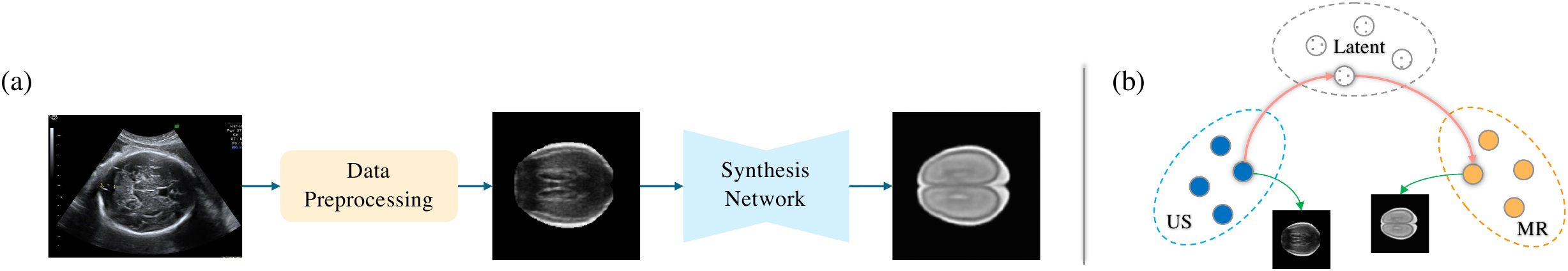}
  \caption{(a) Overview of the proposed framework. Images from left to right: original US, pre-processed US, and the synthesised MR; (b) assumption of the shared latent space.}
  \label{fig:pipe}
\end{figure}

In this paper, we propose an anatomy-aware framework for US-to-MR image synthesis. An overview of the proposed framework is shown in Fig.~\ref{fig:pipe}. Based on an assumption that US and MR modalities share a common latent representation, we design an anatomically constrained learning approach to model the mapping from US to MR images,
that is achieved via an adversarial learning architecture. To the best of our knowledge, this is the first attempt to synthesise MR images from unpaired US images in a self-supervised manner. The proposed method is evaluated both qualitatively and quantitatively, which demonstrate that, even with highly-imbalanced data, neurologically-realistic images can be achieved.

\section{Method}
Given a fetal neurosonography image, our work aims to generate the corresponding MRI-like image. In this section, we present the proposed framework for US-to-MR image synthesis. Specifically, the US image is first pre-processed~\cite{namburete2018fully} to provide spatially normalised data. Then we employ an original data-driven learning-based approach to synthesise the corresponding MR image from the US input. The design of the proposed anatomy-aware synthesis framework is shown in Fig.~\ref{fig:frame}. Given a source US image, the corresponding MR image is synthesised with reference to the real MR image domain.
As our available US and MR data is unpaired, constraints on both pixel-level (\emph{rec. loss}) and feature-level (\emph{dis. loss}) are proposed to ensure anatomical consistency during the synthesis. Referring to Fig.~\ref{fig:frame} during inference/test, only blocks A, B, C and Attention are used. Next, we describe the key steps in the proposed framework in detail.

\subsection{Anatomy-Aware Synthesis}
Paired fetal brain US and MR data is uncommon in clinical practice, and in our case was not available. Therefore, directly learning the mapping from US to MR using conventional deep learning based methods is not applicable to our task. As a result, we propose to model the task as a synthesis framework by enforcing the synthesised MR images to lie in a similar data distribution to real MR images. However, a valid and important constraint is that clinically important anatomical structures should be correctly mapped between the two modalities. Thus we specifically design anatomy-aware constraints to guarantee that the synthesis process is anatomically consistent.

\begin{figure}[t]
  \includegraphics[width=\textwidth]{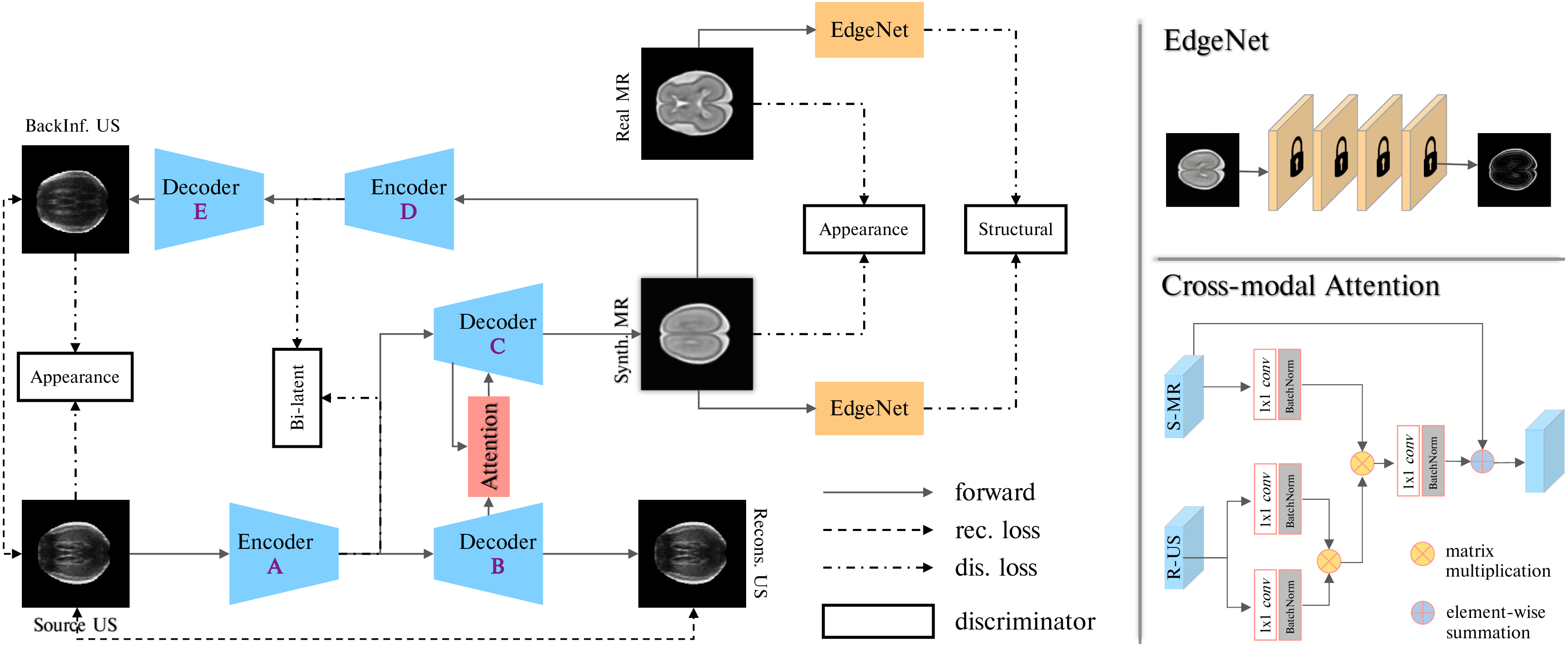}
  \caption{Architecture of the proposed anatomy-aware synthesis framework (the blue block in Fig.~\ref{fig:pipe}(a)). Detailed structures of the \textit{Cross-modal Attention} and \textit{EdgeNet} modules are shown on the right side. The locker symbol indicates a frozen layer.}
  \label{fig:frame}
\end{figure}

\noindent\textbf{Anatomical feature extraction:}
It is rare to have clinical MR and US of the same fetus at the same gestational age, and even if available, the data is not simultaneously co-registered as has often been assumed in other medical image analysis work~\cite{zhao2018towards,nie2017medical}. This makes the modelling of the mapping quite difficult. Without the availability of paired data, we assume the US and MR images share a similar anatomical latent space (Fig.~\ref{fig:pipe}(b)). Based on this assumption, we propose to extract the underlying anatomical features and synthesise the corresponding MR domain data from it, instead of from the original image domain. Specifically, we utilise an autoencoder (encoder-A$\rightarrow$decoder-B) to extract the latent features, as shown in the bottom-left part of the framework (Fig.~\ref{fig:frame}). Assume $\left\{x_U^i\right\}_{i=1}^n$ the set of $n$ original US images where $x_U^i \in \mathcal{X}_U$ is the $i^{th}$ image, the extracted anatomical feature $y^i$ can be formulated as $y^i=F(x_U^i)$ where $F()$ is the extractor.

\noindent\textbf{Bi-directional latent space consistency:}
The extracted anatomical features are fed into a decoder-C to generate a synthesised MR image. Since there is no pixel-level supervision available for \textit{Synth. MR}, we propose to add a backward-inference path (encoder-D$\rightarrow$decoder-E) in the reverse direction that reconstructs the original US image. At the end of the encoder for this reverse path, the encoded feature in the latent space can be extracted. Denoting this encoded feature as $y_b^i$, we propose a bi-directional latent space consistency constraint, based on the anatomical-feature-sharing assumption. This constraint enforces $y^i$ and $y_b^i$ to be similar in latent feature distribution, by a discriminator (\textit{Bi-latent} block in Fig.~\ref{fig:frame}) accompanied with adversarial learning.

\noindent\textbf{Structural consistency:}
The anatomical feature extraction module encodes the main structure of the US image. However, for the MR domain data the image structure is quite different in appearance than in the US domain. To synthesise realistic MR images, we propose to constrain the structures of the \textit{Synth. MR} and the \textit{Real MR} to be similar. Due to the unpaired nature of our task, we enforce the structures to lie in a similar distribution. Specifically, we extract the edges of the \textit{Synth. MR} and \textit{Real MR} by an EdgeNet and measure edge similarity by a structural discriminator (\textit{Structural} block in Fig.~\ref{fig:frame}). The detailed structure of the EdgeNet is illustrated in Fig.~\ref{fig:frame} top-right, which consists of four $3\times3$ convolutional layers with parameters fixed.

\subsection{Cross-modal Attention}
As described to this point, MR image synthesis is mainly guided by the latent features $y^i$ from the end of encoder-A. To provide cross-modal guidance, we propose a cross-modal attention scheme between the US decoder-B and the MR decoder-C, shown as the \emph{Attention} (red) block in Fig.~\ref{fig:frame}. To this end, the US features are reformulated as self-attention guidance for MR image generation, and the guidance is implicitly provided at the feature level. This attention scheme simply consists of several $1\times1$ convolutional layers with a residual connection, which has no influence on the original feature dimension.

We denote the features (R-US in Fig.~\ref{fig:frame}) from the US decoder-B as $f_U$ and the features (S-MR in Fig.~\ref{fig:frame}) from the MR decoder-C as $f_M$, the revised feature after the cross-modal attention scheme can be formulated as: $\widetilde{f}_M=\eta(\delta(f_U)^T \cdot \phi(f_U) \cdot g(f_M))+f_M$, where $\eta,\delta,\phi,g$ are linear embedding functions and can be implemented by $1\times1$~convolutions. By using cross-modal attention, the features do not only consider local information (favoured by CNNs) but also non-local context features from both self- and cross-modal guidance.

\subsection{Joint Adversarial Objective}
Here we formally define the objective functions for model training. As mentioned before and illustrated in Fig.~\ref{fig:frame}, there are two objectives: a pixel-level reconstruction objective and a distribution similarity objective. The pixel-level reconstruction is achieved by an L1-norm, while the distribution similarity is achieved by the discriminator in adversarial learning~\cite{goodfellow2014generative}.

For the forward direction (US-to-MR), we denote the reconstructed US as $\hat{x}_U \in \hat{\mathcal{X}_U}$, the latent feature as $y \in \mathcal{Y}$, the synthesised MR and real MR as $\hat{x}_M \in \hat{\mathcal{X}}_M$ and $x_M \in \mathcal{X}_M$ respectively. The forward objective is defined as:
\begin{equation}
\{\min\mathcal{L}_f~|~\mathcal{L}_f=\lambda\mathcal{L}_{lat}+\mathcal{L}_{app}+\mathcal{L}_{stru}\},
\end{equation}
\begin{equation}
  \mathcal{L}_{lat}=\mathbb{E}_{x_U\in\mathcal{X}_U}\|G_U(F(x_U))-x_U\|_1,
\end{equation}
\begin{equation}
    \mathcal{L}_{app}=\mathbb{E}_{x_M\in\mathcal{X}_M}(D_{app}(x_M))+\mathbb{E}_{y\in\mathcal{Y}}\log(1-D_{app}(G_M(y))),
\end{equation}
\begin{equation}
  \mathcal{L}_{stru}=\mathbb{E}_{x_M\in\mathcal{X}_M}(D_{stru}(E(x_M)))+\mathbb{E}_{y\in\mathcal{Y}}\log(1-D_{app}(E(G_M(y)))).
\end{equation}

Here $G_U$ is the decoder-B used to generate the reconstructed US, $G_M$ is the decoder-C to synthesise the MR, and $\hat{x}_U=G_U(F(x_U))$, $\hat{x}_M=G_M(y)$. $D_{app}$ and $D_{stru}$ are the discriminators (by four \textit{conv} layers) to measure appearance and structure similarity respectively. $E$ represents the EdgeNet. $\lambda$ is a weighting parameter to balance the objective terms and is empirically set to 10.

For the reverse (backward-inference) path, the back-inferred US from the \textit{Synth. MR} is denoted as $\widetilde{x}_U \in \widetilde{\mathcal{X}}_U$ and the back-inferred feature at the end of encoder-D as $y^{back}\in\mathcal{Y}^{back}$, the reverse objective is defined as:
\begin{equation}
  \{\min\mathcal{L}_r~|~\mathcal{L}_r=\lambda\mathcal{L}_{proj}+\mathcal{L}_{app}^{back}+\mathcal{L}_{bi}\},
\end{equation}
\begin{equation}
  \mathcal{L}_{proj}=\mathbb{E}_{\tilde{x}_U\in\tilde{\mathcal{X}}_U,x_U\in\mathcal{X}_U}\|\tilde{x}_U-x_U\|_1,
\end{equation}
\begin{equation}
    \mathcal{L}_{app}^{back}=\mathbb{E}_{x_U\in\mathcal{X}_U}(D_{app}^{back}(x_U))+\mathbb{E}_{y^{back}\in\mathcal{Y}^{back}}\log(1-D_{app}^{back}(G_{BU}(y^{back}))),
\end{equation}
\begin{equation}
  \mathcal{L}_{bi}=\mathbb{E}_{y\in\mathcal{Y}}(D_{bi}(y))+\mathbb{E}_{y^{back}\in\mathcal{Y}^{back}}\log(1-D_{bi}(y^{back})).
\end{equation}

Here $G_{BU}$ is the decoder-E used to back project the US and $\tilde{x}_U=G_{BU}(y^{back})$. $D_{app}^{back}$ and $D_{bi}$ are the discriminators to measure the backward-inference similarity and bi-directional latent space similarity, respectively.
Then the final training model loss based on the above joint adversarial objective functions is:
\begin{equation}
  \mathcal{L}=\mathcal{L}_f+\mathcal{L}_r.
\end{equation}

\section{Experiments}
\subsection{Data and Implementation Details}

We evaluated the proposed synthesis framework on a dataset consisting of healthy fetal brain US and MR volume data. The fetal US data was obtained from a multi-centre, ethnically diverse dataset~\cite{papageorghiou2014international} of 3D ultrasound scans collected from normal pregnancies. We obtained the MR data from the CRL fetal brain atlas~\cite{gholipour2014construction} database and data scanned at Hammersmith hospital. As proof of principle, we selected US and MR data at the gestational age of 23 weeks. In total, we used 107 US volumes and 2 MR volumes, from which approximately 36,000 2D slices were extracted for US and 600 slices for MR. We used 80\% of the total data as the training and validation set, and the remaining 20\% for testing. Our model was implemented by simple \emph{conv, up-conv}, and \emph{max-pooling} layers. Skip connections were added between each encoder-decoder pair to preserve structural information. An Nvidia Titan V GPU was utilised for model training. The complete model was trained end-to-end. The testing phase only takes an US scan as input without any discriminators and the reverse path.

\subsection{Evaluation Metrics}\label{sec:metric}
Since we are not using US-MR paired data, traditional evaluation metrics like PSNR (Peak Signal-to-Noise Ratio) and SSIM (Structural Similarity) cannot be applied. Therefore, we evaluated the quality of the synthesised MRI using two alternative metrics: 1) the Mean Opinion Score (MOS) and 2) a Jacobian-based registration metric. The MOS is expressed in a rating range between 1 and 5, in which 5 indicates \emph{excellent} while 1 \emph{bad}. The MOS test was performed by two groups (2 medical experts and 11 beginners) with 80 samples shown to each participant. For the registration-based objective score, we performed a deformable registration (FFD~\cite{rueckert1999nonrigid}) between the synthesised MR and the real MR at a similar imaging plane, and then computed the average Jacobian of the deformation (normalised to [0,1]) required to achieve this. We assume a high-quality synthesised MRI will have a lower Jacobian for the registration.

\begin{figure}[t]
  \centering
  \includegraphics[width=0.96\textwidth]{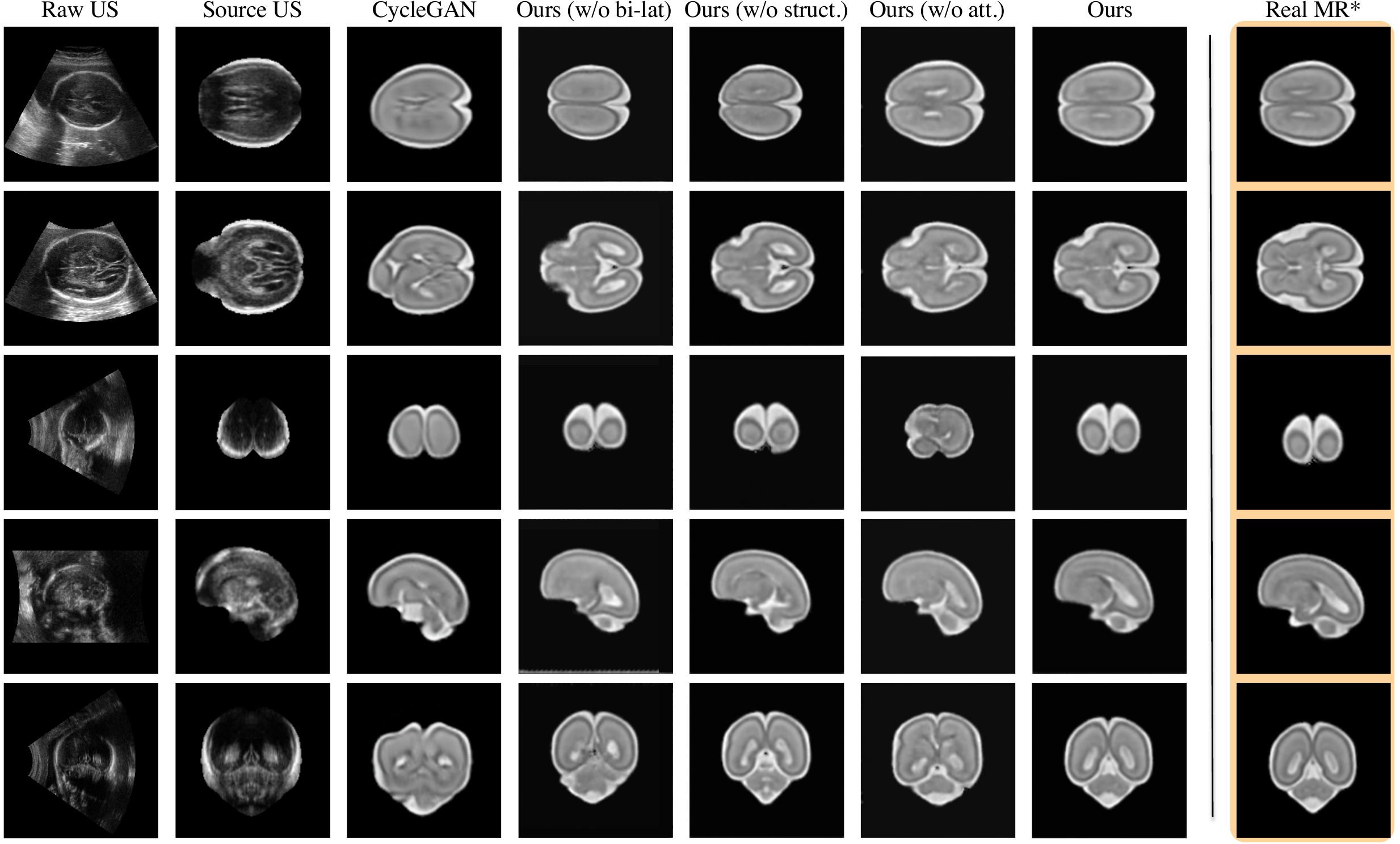}
  \caption{Qualitative performance on the US-to-MR image synthesis. Each row shows an example sample. From left to right are the original raw US, pre-processed US, synthesised MR by CycleGAN~\cite{CycleGAN2017} and Ours (with its counterparts), and the reference real MR. *Note that the last column is \textbf{NOT} the exact corresponding MR images.}
  \label{fig:vis}
\end{figure}

\subsection{Results}
We present synthesised MR results and image synthesis quality evaluations. In the testing phase, all the discriminators were removed, leaving only the anatomical feature extractor and the synthesis decoder, which took the test US image as input. Several synthesised MR results with the corresponding US input samples are shown in Fig.~\ref{fig:vis}. Note that the last column shows a real MR example for comparison (\textbf{not} in direct correspondence to the synthesised one). From the results, we observe that the visual appearance of the synthetic MR images is very similar to the real ones, and is visually superior to the results from CycleGAN~\cite{CycleGAN2017}. In addition, anatomical structures are well preserved between the source US and the synthetic MR image in each case.

\begin{table}[h!]
  \caption{Quantitative evaluation on our synthesised MR images for MOS test (on both experts and beginners) and deformation score, with the comparison to several possible solutions. MOS the higher the better while deformation the lower the better.}
  \label{tab:mos}
  \begin{adjustbox}{max width=\textwidth}
  \begin{tabular}{ll|ccc|ccc|cc}
    \toprule
    \multicolumn{2}{l|}{Method} & AE & GAN & CycleGAN & Ours (w/o bi-lat) & Ours (w/o struct.) & Ours (w/o att.) & Ours & Real\\
    \midrule
    & Expert & 1.00 & 2.05 & 2.50 & 3.05 & 3.45 & 3.30 & \textbf{3.90} & \textcolor{gray}{4.35}\\
    \multirow{-2.25}{*}{\begin{sideways} MOS$\uparrow$\end{sideways}} & Beginner & 1.01 & 2.75 & 3.42 & 3.69 & 3.87 & 3.65 & \textbf{4.08} & \textcolor{gray}{4.23}\\
    \midrule
    \multicolumn{2}{l|}{Deformation$\downarrow$} & 0.97 & 0.78 & 0.66 & 0.55 & 0.65 & 0.47 & \textbf{0.46} & \textcolor{gray}{0.00} \\
    \bottomrule
  \end{tabular}
\end{adjustbox}
\end{table}

Quantitative results are reported in Table~\ref{tab:mos}. We compare our method with the vanilla
autoencoder (\emph{AE}), GAN~\cite{goodfellow2014generative}, and CycleGAN~\cite{CycleGAN2017}. We also performed an ablation study by removing the bi-directional latent consistency (\emph{w/o bi-lat}) module, removing the structural consistency module (\emph{w/o struct.}), or removing the cross-modal attention module (\emph{w/o att.}). The results in Table~\ref{tab:mos} suggest that the proposed method performs better than the other possible solutions, and also supports the inclusion of each proposed term in our model.

\section{Conclusion}
In this paper, we have presented to our knowledge the first attempt to synthesise MRI-like fetal brain data from unpaired US data, by a new anatomy-aware self-supervised framework. Specifically, we first extract shared features between the two imaging modalities and then synthesise the target MRI by a group of anatomy-aware constraints. A cross-modal attention scheme was introduced to incorporate non-local guidance across the different modalities. Experimental results demonstrated the proposed framework effectiveness both qualitatively and quantitatively, with comparison to alternative architectures. We believe the proposed method may be useful within analysis tasks such as US-MR alignment and for communicating US findings to paediatricians and patients. While we made the first step for US-to-MR synthesis in 2D in this paper, the extension to 3D synthesis would be an interesting direction to explore in future work.

\subsubsection*{Acknowledgments.}
We thank Andrew Zisserman for many helpful discussions, the volunteers for assessing images, NVIDIA Corporation for a GPU donation, and acknowledge the ERC (ERC-ADG-2015 694581), the EPSRC (EP/M013774/1, EP/R013853/1),
the Royal Academy of Engineering Research Fellowship programme and the NIHR Biomedical Research Centre funding scheme.

%
\bibliographystyle{splncs04}
\bibliography{mybib_short}
\end{document}